\documentclass[review]{elsarticle}
\usepackage{lineno,hyperref}
\usepackage{graphicx}
\usepackage{pdfpages}
\usepackage{times}
\usepackage{epsfig}
\usepackage{amsmath,amssymb,amstext}
\usepackage{amssymb}
\usepackage{algorithm}
\usepackage{algorithmic}

\usepackage{booktabs}
\usepackage{epstopdf}
\usepackage{subfigure}
\usepackage{booktabs}
\usepackage{diagbox}
\usepackage{float}
\usepackage{makecell}
\usepackage{tabularx}
\usepackage{amsthm}

\usepackage{mathrsfs} 
\usepackage{hyperref}
\usepackage{lipsum}
\usepackage{amssymb}
\usepackage{amsmath}
\usepackage{mathrsfs} 
\usepackage{bbm}
\usepackage{booktabs}
\usepackage{threeparttable}
\usepackage{amsfonts}
\usepackage{multirow}
\usepackage{stmaryrd}
\modulolinenumbers[1]
\usepackage{lineno}

\begin{document}

\begin{frontmatter}
\title{Adaptive Pseudo-Siamese Policy Network for Temporal Knowledge Prediction}

\author{Pengpeng Shao$^{1,2}$, Tong Liu$^1$, Feihu Che$^{1,2}$, Dawei Zhang$^1$, Jianhua Tao$^{1,2,3}$}

\address{$^1$National Laboratory of Pattern Recognition, Institute of Automation, Chinese Academy of Sciences, Beijing, China; \\$^2$School of Artificial Intelligence, University of Chinese Academy of Sciences, Beijing, China;\\ $^3$CAS Center for Excellence in Brain Science and Intelligence Technology, Beijing, China\\}

\begin{abstract}
Temporal knowledge prediction is a crucial task for the event early warning that has gained increasing attention in recent years, which aims to predict the future facts by using relevant historical facts on the temporal knowledge graphs. There are two main difficulties in this prediction task. First, from the historical facts point of view, how to model the evolutionary patterns of the facts to predict the query accurately. Second, from the query perspective, how to handle the two cases where the query contains seen and unseen entities in a unified framework. Driven by the two problems, we propose a novel adaptive pseudo-siamese policy network for temporal knowledge prediction based on reinforcement learning. Specifically, we design the policy network in our model as a pseudo-siamese policy network that consists of two sub-policy networks. In sub-policy network I, the agent searches for the answer for the query along the entity-relation paths to capture the static evolutionary patterns. And in sub-policy network II, the agent searches for the answer for the query along the relation-time paths to deal with unseen entities. Moreover, we develop a temporal relation encoder to capture the temporal evolutionary patterns. Finally, we design a gating mechanism to adaptively integrate the results of the two sub-policy networks to help the agent focus on the destination answer. To assess our model performance, we conduct link prediction on four benchmark datasets, the experimental results demonstrate that our method obtains considerable performance compared with existing methods.
\end{abstract}


\end{frontmatter}


\section{Introduction}
\label{sec::introduction}

Knowledge graphs (KGs), which store plenty of static triple facts in form of (subject, relation, object), have been widely used in many natural language processing applications, such as question answering~\cite{QA}, recommendation systems~\cite{RecommenderSystems}, retrieval systems~\cite{InformationRetrieval}. However, with the rise of triple facts, many facts present dynamic attributes, and they only hold in a specific period or a certain point in time, thus temporal knowledge graphs (TKGs) and corresponding temporal knowledge reasoning tasks receive increasingly extensive attention and research in recent years. 

TKGs reasoning has two forms - interpolation and extrapolation. On a TKGs with timestamps varying from $[\emph{t} _{0}, \emph{t} _{T}]$, interpolation TKGs reasoning aims to infer the answer for the query with time $\emph{t} \in [\emph{t}_{0}, \emph{t}_{T}]$. While extrapolation setting only employs historical knowledge to predict future facts for time $\emph{t} $ ($\emph{t}  > \emph{t} _{T}$), thus we call the extrapolation setting temporal knowledge prediction here. Furthermore, the temporal knowledge prediction also is a crucial task on event early warning. In this work, we focus on dealing with the problems in temporal knowledge prediction.

Recently, many temporal knowledge prediction methods~\cite{CyGNet, DyRep, Renet} based on embedding are presented, and only a few works~\cite{CluSTeR, TITer} employ the models based on reinforcement learning (RL) to predict the temporal knowledge. Although embedding-based approaches are convenient to model the knowledge and obtain considerable performance, they fail to take the symbolic compositionality of KG relations into consideration, which limits their application in more complex reasoning tasks. While RL-based approaches can allow the agent to find the answer to the query by traversing the path in the interaction with the complex environment, thus they possess powerful adaptability. However, there exist two main difficulties for the agent in the process of traversing. First, how to model the evolutionary patterns of the historical facts to predict the future facts accurately. Second, the queries are usually uncertain, and it usually contains seen entity or unseen entity, thus the second difficulty is how to handle the two cases where the query contains seen and unseen entities in a unified framework. Many related works have been proposed to handle these two problems. In the embedding-based approaches, to model the temporal pattern of the facts, Know-Evolve~\cite{Know-Evolve} and its extension DyRep~\cite{DyRep} model the occurrence of the facts as temporal point process, Re-NET~\cite{Renet} models the occurrence of a fact as a probability distribution conditioned on temporal sequences of past knowledge graphs. To copy with the second obstacle, XERTE~\cite{XERTE} develops a temporal relational attention mechanism and predicts over the relevant substructure of TKGs by propagating attention, TITer~\cite{TITer} uses the inductive mean representation of the trained entities with the same co-occurrence query relation to represent unseen entities, thus obtains better entity embedding distribution and answer score distribution. 
However, the unseen entity is an entity that does not exist in historical facts and has no historical neighbours. Thus, for the query containing the unseen entity, XERTE~\cite{XERTE} can not propagate attention to entities of previous facts, TITer~\cite{TITer} fails to select candidate temporal actions to search for the answer.

In this work, we propose a novel RL-based temporal knowledge prediction model to address the two problems. Specifically, we design the policy network of the proposed model as a pseudo-siamese policy network that consists of two sub-policy networks. The sub-policy network I is designed to capture the static evolutionary patterns, and the agent I can search for the answer for the query along the entity-relation paths instead of the time information. In sub-policy network II, to deal with unseen entities, we add semantic edges in the TKGs, the agent II searches for the answer for the query along the relation-time paths instead of the entity information. Furthermore, we develop a temporal relation encoder to capture the temporal evolutionary patterns. Finally, in order to make the two sub-policy networks compensate for each other, we employ a gating mechanism to adaptively integrate the results of two sub-policy networks to help the agent focus on the destination entity. The extensive experimental results show that our method obtains considerable performance compared with existing approaches and demonstrate the effectiveness and superiority of our method.
Our contributions are as follows:
\begin{itemize}
\item We develop a novel RL-based temporal knowledge prediction approach to deal with the two cases where the query contains seen and unseen entities in a unified framework.
\item We advocate the importance of more comprehensive modelling of the evolutionary patterns of the facts, thus develop one sub-policy network to capture static evolutionary patterns, and design another sub-policy network and temporal relation encoder to model temporal evolutionary patterns. Finally, we employ a gating mechanism to adaptively integrate the results of two sub-policy networks.  
\item We propose a new type of edge to build the relationship between the query containing the unseen entity and historical facts and design a novel method to handle the special type of query.
\item Experimental studies on four TKGs datasets demonstrate that our method achieves state-of-the-art performance.
\end{itemize}

\section{Related Work} 

\subsection{Static KGs Reasoning}

The static KGs reasoning methods can be roughly grouped into three categories: embedding-based approaches, RL-based approaches, and logic rule-based approaches. Therein, embedding-based approaches are the most popular one by virtue of their high efficiency and outstanding effectiveness, which aim to project the entities and relations in KGs into vector space and represent them as low-dimensional embeddings. This type of method is also broadly classified into three paradigms: (i) Translational distance-based models~\cite{TransE, TransD, TransH}. (ii) Tensor factorization-based models~\cite{RESCAL, DisMult, ComplEx, Tucker}. (iii) Neural network-based models~\cite{ConvE, R-GCN, CompGCN}.
 
Although embedding-based approaches present simplicity and convenience in modelling knowledge and obtain considerable performance, they are less sensitive to the reasoning distance and ignore the logical rules between relations and paths, which limits their application in more complex reasoning tasks and lacks interpretability. While RL-based approaches can allow the agent to find the answer to the query by traversing the path on KGs, which enables it to learn reasoning rules from relation paths. DeepPath~\cite{DeepPath} is the first multi-hop reasoning work based on RL, it targets at searching for generic representative paths between pairs of entities. MINERVA~\cite{MINERVA} utilizes the history path to facilitate the agent searching answer entities of a particular KG query in an end-to-end fashion. Based on MINERVA, M-Walk~\cite{M-Walk}  and Multi-HopKG~\cite{MultiHopKG} adopt Monte Carlo tree search and pre-trained embedding model to overcome the problem of sparse rewards, respectively. However, all these methods can not model the evolutionary patterns of the facts in TKGs.

\subsection{Temporal KGs Reasoning}
According to the relationship between query time and maximum training time, TKGs reasoning can broadly be classified into two forms - interpolation and extrapolation. Interpolation reasoning~\cite{tTransE, HyTE, TADistMult, DESimplE, TNTComplex} aims to infer new facts at history timestamps by employing historical information and future information.
Corresponding extrapolation reasoning aims to predict future facts only based on historical information, so we call this reasoning task temporal knowledge prediction here. In view of its great practical value, especially in the event early warning, many temporal knowledge prediction works have been presented in recent years. Know-Evolve~\cite{Know-Evolve} and its extension DyRep~\cite{DyRep} employ the temporal point process to model the occurrence of the temporal facts in TKGs. While Re-NET~\cite{Renet} models the occurrence of the temporal fact as a probability distribution conditioned on temporal sequences of past knowledge graphs.
To model repetitive facts, CyGNet~\cite{CyGNet} designs a copy mode to learn  from the known facts that appeared in history. XERTE~\cite{XERTE} develops a temporal relational attention mechanism and predicts the future facts by propagating attention over the relevant substructure of TKGs. Especially, CluSTeR~\cite{CluSTeR} and TITer~\cite{TITer} are two temporal knowledge prediction approaches based on RL. CluSTeR~\cite{CluSTeR} first employs the agent to induce relevant clues from historical facts, then adopts an embedding model to deduce answers from the obtained clues. While TITer~\cite{TITer} forces the agent to travel on historical knowledge graph snapshots and find the answer for the query directly. Furthermore, to deal with the unseen entities in the query, TITer presents an embedding method to obtain a more reasonable initial embedding for unseen entities, thus improving the inductive inference ability of the model. However, TITer~\cite{TITer} only uses the inductive mean representation to obtain better entity embedding distribution and answer score distribution, and fails to select candidate temporal edges to search for the answer except for the self-loop edges for the query with an unseen entity.

\section{The Proposed Model}

\subsection{Notations and Problem Formulation}
The temporal facts in TKGs is defined by quadruples $(s, r, o, t)$ $\subseteq$  $\mathcal{E}$ $\times$ $\mathcal{R}$ $\times$ $\mathcal{E}$ $\times$ $\mathcal{T}$, where $\mathcal{E}$, $\mathcal{R}$, and $\mathcal{T}$ denote a finite set of entities, relations, and timestamps, respectively. And TKGs also can be represented as graph snapshots over time, namely $\mathcal{G}$ = $\{\mathcal{G}_{t_1}, \mathcal{G}_{t_2}, \dots,\mathcal{G}_{t_T}\}$ = $\{(s, r, o, t_{l})\}_{t_{l}=1}^{\vert\mathcal{T}\vert}$,  where $s$, $o$ $\in$ $\mathcal{E}$, $r$ $\in$ $\mathcal{R}$, and each $\mathcal{G}_{t_l}$ is a static multi-relational KGs at time $t_l$. Thus TKGs is a collection of event fact at different timestamps, where the timestamps are arranged in ascending order.

The task of temporal knowledge prediction is to predict the missing entity for the query $(s_q, r_q, ?, t_q)$ or $(?, r_q, o_q, t_q)$ according to the historical facts $\mathcal{G}_{0:t_q-1}$. The inverse facts are added into the training dataset, and predicting the subject entity for $(?, r_q, o_q, t_q)$ can be transformed into predicting the object entity for $(o_q, r_q^{-1}, ?, t_q)$ without loss of generality. Note that, not all historical facts are relevant to the query in $\mathcal{G}_{0:t_q-1}$, thus the historical facts that related to the query subject $s_q$ are represented as  $E_{s}$ = $\{(s_q, r, o, t_l) \in \mathcal{G}_{t_l}\vert t_l < t_q\}$, which have same subject entity with query. In addition, we denote the historical facts which related to the query relation $r_q$ as  $E_{r}$ = $\{(s, r_q, o, t_l) \in \mathcal{G}_{t_l}\vert t_l < t_q\}$, that is, the facts in $E_{r}$ have same relation with query. Due to the relation of an entity reflects the roles of the entity, so they have similar semantic meanings. To be more specific, our task is to employ these historical facts that related to the query to answer the given query.

\subsection{Reinforcement Learning System}
The temporal knowledge prediction task is formulated as a finite horizon sequential decision-making problem, and the reinforcement learning system are described as a deterministic Partially Observed Markov Decision Process (POMDP). The system consists of agent and environment. The agent starts from the query entity $s_q$, follows a relation path in $\mathcal{G}$ according to its policy, and stops at the entity which is regarded as the right answer to the query. The environment is described in detail as follows.
\\

\noindent \textbf{States.} Each state can be represented as $s_k = (e_k, t_k, s_q, r_q, o_q, t_q) \in \mathcal{S}$, where $\mathcal{S}$ is the state space, $e_k$ is the entity visited at step $k$, $t_k$ is the timestamp of the action taken at the previous step, and $(s_q, r_q, t_q)$ can be seen as the global context shared by all the states for the given query, and $o_q$ is the answer. Due to the agent starts from the query node $s_q$ and the default initial action is self-loop action, so the initial state is $s_0 = (s_q, t_q, s_q, r_q, o_q, t_q)$.\\

\begin{figure*}
\hspace*{-1.3cm}
\setlength{\abovecaptionskip}{-0.9cm}
\setlength{\belowcaptionskip}{-0.cm}  
\centering
\includegraphics[height=4.1cm, width=14cm]{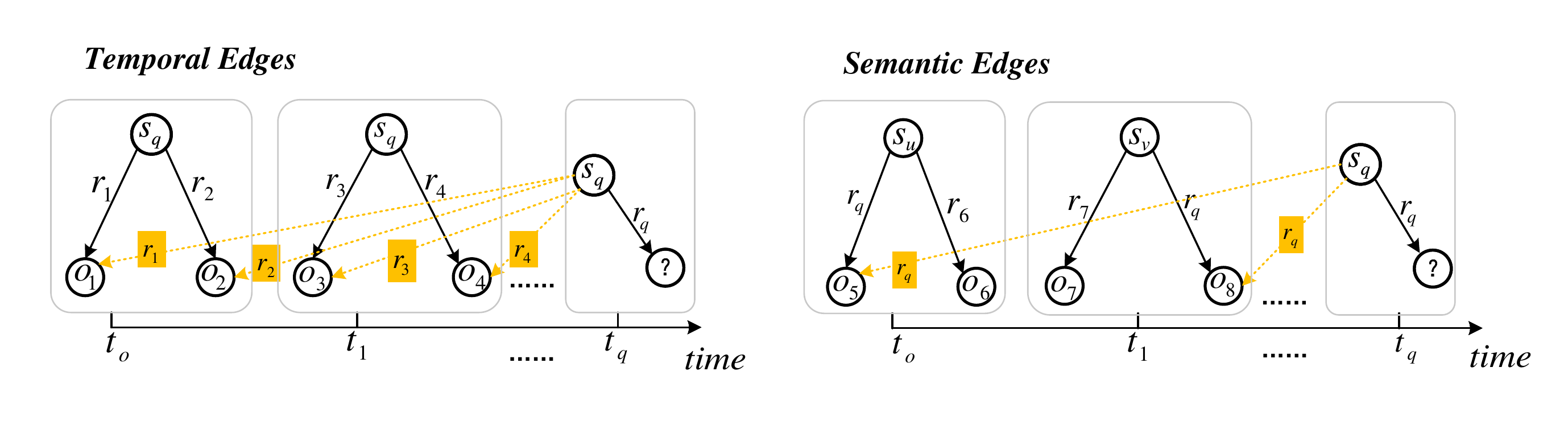} 
\caption{Illustration of the temporal edges and semantic edges} 
\label{fig::figure_actions}
\end{figure*}

\noindent \textbf{Observations.} In the process of searching for answers, the agent can not observe all the states of the environment but the current location and the query information, and the answer $o_q$ remians invisible. Therefore, the observation function is defined as:  $\mathcal{O}((e_k, t_k, s_q, r_q, o_q, t_q)) = (e_k, t_k, s_q, r_q, t_q)$. \\
 
\noindent \textbf{Actions.} Given the current state $(e_{k-1}, t_{k-1}, s_q, r_q, o_q, t_q)$, the set of the possible actions  $\mathcal{A}_k \in \mathcal{A}$ at step $k$ consists of the outgoing edges of $e_{k-1}$, where $\mathcal{A}$ is the action space.
Here, we describe three types of outgoing edges of entity $e_{k-1}$ at step $k$ as follows:
(i) Self-loop Edges. The self-loop edges (i.e. $(e_{k-1}, r_{self}, e_{k-1}, t_{k-1})$) not only allow the agent to stay in a place but also allow it to stop adaptively when it searches for a certain number of steps or argues that it has found the final answer. 
(ii) Temporal Edges. If there exist related historical facts $(e_k, r, o, t_l)$ for the current entity $e_k$ at time $t_k$, we build the temporal edge between $e_k^{t_k}$ and $o^{t_l}$ through relation $r$. For example, as shown in the left of Figure~\ref{fig::figure_actions}, in the initial state, when the query contains the seen entity $s_q$ and we can find the related historical facts $(s_q, r_1, o_1, t_0)$, we build the temporal edges $(s_q^{t_q}, r_1, o_1, t_0)$ .
(iii) Semantic Edges. When the query contains the unseen entity $s_q$, and there are no exist the related historical facts $(s_q, r, o, t_l)$ for the entity $s_q$. Thus, in this case, the agent can only stay in a place through the self-loop edges. Fortunately, in the initial state, there exist the historical facts $(s_v, r_q, o, t_l)$ that related to the query relation $r_q$, as is $(s_v, r_q, o_8, t_1)$ shown in the right of Figure~\ref{fig::figure_actions}, thus we build the semantic edges $(s_q, r_q, o_8, t_1)$. Note that, the semantic edges only exist in the first step to bridge the gap between the query and historical facts. To summarize, when the query contains seen entity, the agent can walks along the above three types of edges in the first step, then walks along temporal edges or self-loop edges to find answers. When the query contains unseen entity, the agent only can walks along the semantic edges in the first step to arrive at seen entity, and then walks along temporal edges or self-loop edges to find answers. Therefore, $\mathcal{A}_0 = \{(r_{self}, s_q, t_q) \cup \{(r', e', t')| (s_q, r', e', t') \in \mathcal{G}_{0:t_q-1}, t' < t_q \}  \cup (r_q, e', t')| (s_u, r_q, e', t') \in \mathcal{G}_{0:t_q-1}, (s_q, r', e', t') \notin \mathcal{G}_{0:t_q-1}, t' < t_q \}$. And $\mathcal{A}_k = \{(r_{self}, e_k, t_k)\} \cup\{(r', e', t')| (e_k, r', e', t') \in \mathcal{G}_{0:t_k-1}, t' < t_k \} $, $k >0$.
Moreover, given the action $(r, e, t)$, here, we denote $(r, e)$ and $(r, t)$ as static action and temporal action, respectively.\\

\noindent \textbf{Transition.} Given the current location $s_{k-1}$, once the action $a_{k}$ is determined, the current location $s_{k-1}$ is transferred to the next location $s_{k}$. Here, the transition function $\delta$ : $\mathcal{S} \times \mathcal{A} \to \mathcal{S}$ defined by $\delta(s_{k-1}, a_{k}) = s_{k} = (e_{k}, t_{k}, s_q, r_q, o_q, t_q)$. Therein, the query information and answer are unchanged, the action $a_{k} = (r_{k}, e_{k}, t_{k})$.\\

\noindent \textbf{Rewards.} After $K$-hop navigation, the agent reaches its final state $(e_{K}, t_{K}, s_q, r_q, o_q, t_q)$. The agent receives a terminal
reward of 1 if $e_{K} = o_q$, and 0 otherwise. Formally, the reward mechanism is defined as : $R(s_K) = \mathbb{I}(e_{K} == o_{q})$.

\subsection{Policy Network}
\begin{figure}[H]
\hspace*{-1.2cm}
\setlength{\abovecaptionskip}{-0.9cm}
\setlength{\belowcaptionskip}{-0.cm}  
\centering
\includegraphics[height=6.3cm, width=13.5cm]{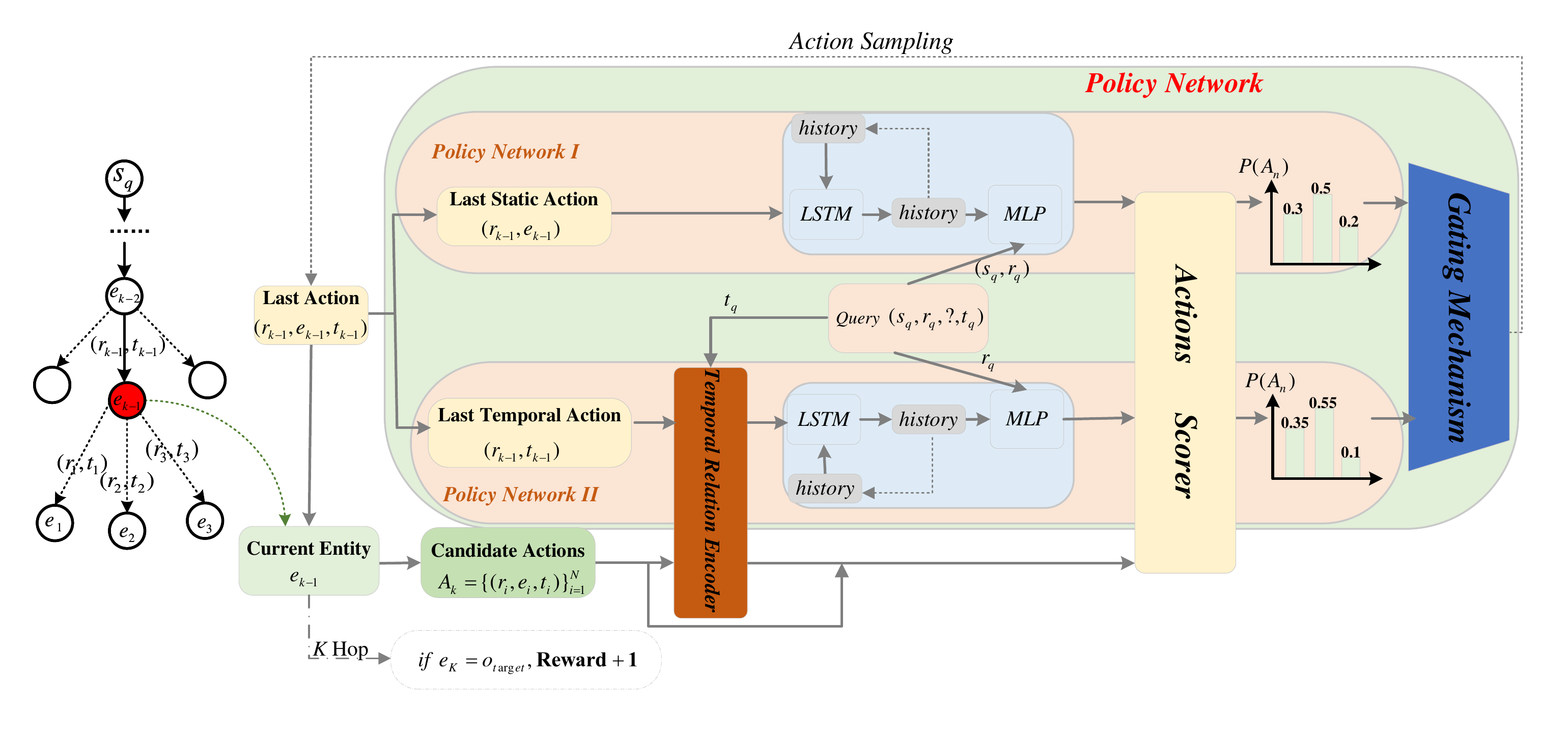} 
\caption{An overview of our model. Given the query $(s_q, r_q, ?, t_q)$, the agent starts from query subject $s_q$, then samples an outgoing edge (action) to walk and reaches the next entity according the policy network $\pi_{\theta}$. For example, after $k$-1 hop navigation, the agent arrive at $e_{k-1}$ (red node), and we  can also obtain a historical path consisting of $k$-1 historical actions. Then we use two sub-policy networks to encode the two variant path of this historical path and obtain the scores of each candidate action, respectively. Finally, we employ gating mechanism to obtain the final scores of each candidate action adaptively and sample the next action.}
\label{fig::figure_framework}
\end{figure}

The policy network $\pi_{\theta}$ is parameterized by using the action path history and global context (query information), and it models the action of the agent in a continuous space, thus we need to calculate the similarity between the continuous value outputted by the policy network and the candidate actions to determine the probability of selecting each candidate action. Here, to deal with the case of the seen and unseen entities simultaneously in a unified framework and model the evolutionary patterns of the facts, we design two similar sub-policy networks to encode the path and score the candidate actions, respectively, then employ a gating mechanism to obtain a final score of each candidate action. The overview of our model is presented in Figure~\ref{fig::figure_framework}.

\subsubsection{Policy Network I} The search history consisting of $k$ actions is $h_k = \big((s_q, t_q), (r_{self}, s_q, t_q), (r_1, e_1, t_1),\\
..., (r_k, e_k, t_k)\big)$. To deal with the general case that the query contains seen entity, we use the policy network to encode the action path history $h_k$ as usual. However, to model the static and temporal evolutionary patterns, here, we use one policy network (policy network I) to encode the static action path history $h_k^s = \big((s_q, t_q), (r_{self}, s_q), ..., (r_k, e_k)\big)$, and employ another policy network (policy network II) to encode the temporal action path $h_k^t = \big((s_q, t_q), (r_{self}, t_q), (r_1, t_1), ..., (r_k, t_k)\big)$. With this design, we can model both static and temporal evolutionary patterns in the general case that the query contains seen entity. The static action path $h_k^s$ is encoded by policy network I as follows, 
\begin{eqnarray}
\begin{aligned}
\label{eq::encodehistory1}
&h_0^{s} =  \emph{\rm LSTM}(0, [r_{self}, s_q]) \\
&h_1^{s} =  \emph{\rm LSTM}(h_0, [r_1, e_1]) \\
&~~~~~~~~~~~~~~\cdots \\
&h_k^{s} =  \emph{\rm LSTM}(h_{k-1}, [r_k, e_k])
\end{aligned}
\end{eqnarray} 

We then calculate probability of selecting each candidate action in policy network I. First, we denote $N$ candidate actions at step $k$ as  $\emph{\rm A}_k = \{(r_i, e_i, t_i)\}_{i=1}^{N}$. Similarly, the static candidate actions is $\emph{\rm A}_k^{s} = \{(r_i, e_i)\}_{i=1}^{N}$, and the temporal candidate actions is represented as $\emph{\rm A}_k^{t} = \{(r_i, t_i)\}_{i=1}^{N}$. Thus, the probability distribution $\phi^{s}_{\theta}(a_k|s_{k-1})$ of the candidate actions in the state $s_{k-1}$ are calculated as,
\begin{eqnarray}
\begin{aligned}
\label{eq::probability1} 
\phi_{\theta}^{s}(a_k|s_{k-1}) = \emph{\rm A}_k^{s} \emph{\rm W}_{2}^{s}\emph{\rm ReLU}( \emph{\rm W}_{1}^{s}[h_{k-1}^{s}, s_q, r_q])
\end{aligned} 
\end{eqnarray}
where $\emph{\rm W}_{1}^{s}$ and $\emph{\rm W}_{2}^{s}$ are the training parameters in MLP.

\subsubsection{Policy Network II } As mentioned above that the policy network II is used to encode the temporal action path history $h_k^t = \big((s_q, t_q), (r_{self}, t_q), (r_1, t_1), ..., (r_k, t_k)\big)$ to model the temporal evolutionary patterns in the case that the query contains seen entity. Here, it is worth noting that when the query contains the unseen entity, we bridge the gap between the query and historical facts through the semantic edges defined by query relation, thus we need to design a policy network to encode the temporal action path history $h_k^t = \big((s_q, t_q), (r_{self}, t_q), (r_1, t_1), ..., (r_k, t_k)\big)$ instead of entity information. Therefore, encoding the temporal action path $h_k^t$ not only can model the evolutionary patterns in the general case but also deal with the special case that the query contains the unseen entity.

Here, As the time difference between the facts can better capture temporal evolutionary patterns and express the dynamics of the facts, thus we first employ temporal relation encoder to encode $h_k^t$ to obtain the temporal relation path$\big((s_q, t_q), r_{self}^{t_q}, r_1^{t_1}, ...,\\
 r_k^{t_k}\big)$, then use policy network II to encode the obtained temporal relation path. Formally, the temporal action $r_{k-1}^{t}$ is encoded by temporal relation encoder as follows,
\begin{eqnarray}
\begin{aligned}
\label{eq::TemporalRelationEncoder}
r_{k}^{t} = (\sigma (\emph{\rm W} \Delta_{t}) + \emph{b}) * r_{k}
\end{aligned} 
\end{eqnarray}
Where $w$ and $b$ are the training parameters, $\sigma$ is activation function. ${\Delta}_{t}$ is the timestamps difference between temporal facts, and $\Delta_{t} = t_q - t_k$. Likewise, we learn the temporal relation $\emph{\rm A}_k^{tr}= \{r_i^{t_i}\}_{i=1}^{N}$ of the candidate temporal actions $\emph{\rm A}_k^{t}$ for scoring the actions in a unified form.

We then use policy network II to encode the temporal relation path,
\begin{eqnarray}
\begin{aligned}
\label{eq::encodehistory2}
&h_0^{t} = \emph{\rm LSTM}(0, r_{self}^{t_q})\\
&h_1^{t} =  \emph{\rm LSTM}(h_0, r_1^{t_1})\\
&~~~~~~~~~~~~~~\cdots\\
&h_k^{t} =  \emph{\rm LSTM}(h_{k-1}, r_k^{t_k})
\end{aligned} 
\end{eqnarray}
Afterward, the probability distribution $\phi_{\theta}^{t}(a_k|s_{k-1})$ of the candidate actions in the state $s_{k-1}$ are calculated as,
\begin{eqnarray}
\begin{aligned}
\label{eq::probability2} 
\phi_{\theta}^{t}(a_k|s_{k-1}) = \emph{\rm A}_k^{tr} \emph{\rm W}_{2}^{t}\emph{\rm ReLU}(\emph{\rm W}_{1}^{t}[h_{k-1}^{t}, r_q])
\end{aligned} 
\end{eqnarray}
where $\emph{\rm W}_{1}^{t}$ and $\emph{\rm W}_{2}^{t}$ are the training parameters in MLP.

\subsubsection{Action scorer with gating mechanism}

The future query events are usually uncertain, and it usually contains seen entity or unseen entity. To deal with these two cases in a unified RL framework and capture the evolutionary patterns of the facts more fully in the general case that the query contains seen entity to predict the query accurately, we design such a novel model with two sub-policy networks. Therefore, given two probability distributions of the candidate actions outputted by two sub-policy networks, we design a gating mechanism to adaptively integrate them to help the agent focus on the destination entity and obtain the final probability distribution $\pi_{\theta}(a_k|s_{k-1})$ of  the candidate actions,
\begin{eqnarray}
\begin{aligned}
\label{eq::probability} 
\pi_{\theta}(a_k|s_{k-1}) =\emph{\rm softmax}\big( (1 -  g_t) \ast \phi_{\theta}^{s}(a_k|s_{k-1}) 
+ g_t \ast \phi_{\theta}^{t}(a_k|s_{k-1}) \big)
\end{aligned} 
\end{eqnarray}
\begin{eqnarray}
\begin{aligned}
\label{eq::gatingmechanism} 
g_t =  \emph{\rm sigmoid}(\emph{\rm W}_{g}[h_{k-1}^{t}, r_k^{t_k}, r_q])
\end{aligned} 
\end{eqnarray}
where the gate function is parameterized by the temporal history, the candidate temporal actions, and the query relation. $\emph{\rm W}_{g}$ is the learnable parameter.

\subsection{Optimization}

We set the length of the search path to $K$, and the $K$-hop action path generated by policy network $\pi_{\theta}$ is $\{a_1, a_2, \cdots, a_K\}$. Thus, the policy network is optimized by maximizing the expected reward over all queries in the training set $\mathcal{D}_{train}$,
\begin{eqnarray}
\begin{aligned}
\label{eq::Optimization} 
J(\theta) &=  \mathbb{E}_{(s_q, r_q, o_q, t_q) \sim \mathcal{D}_{train}}[\mathbb{E}_{a_1, \cdots, a_K \sim \pi_{\theta}} [R(s_K|s_q, r_q, t_q)]]
\end{aligned} 
\end{eqnarray}
We then employ the policy gradient method to optimize the policy network.  Specifically, Equation~\ref{eq::Optimization} is optimized by the REINFORCE algorithm, which will iterate through all quadruple in $\mathcal{D}_{train}$ and update $\theta$ with the following stochastic gradient,
\begin{eqnarray}
\begin{aligned}
\label{eq::Optimization2} 
\nabla_{\theta}J(\theta)  \thickapprox \nabla_{\theta} \sum_{k} R(s_K|s_q, r_q, t_q) \emph{\rm log}\pi_{\theta}(a_k|s_{k-1})
\end{aligned} 
\end{eqnarray}

\section{Experiment Result}
\renewcommand\arraystretch{0.2} 
\begin{table}[H] 
\centering 
\small
\setlength{\tabcolsep}{3pt}{
\begin{tabular}{l|l|l|l|l|l|l|l}  
Datasets & $N_{\emph{entity}}$ & $N_{\emph{relation}}$ & $N_{\emph{timestamps}}$ &\emph{Time granularity} & $N_{\emph{train}}$ & $N_{\emph{valid}}$ & $N_{\emph{test}}$\\ \hline  
\makecell[c]{ICEWS2014} &  \makecell[c]{7,128} &  \makecell[c]{230} & \makecell[c]{365} &  \makecell[c]{24 hours}&  \makecell[c]{63685} &  \makecell[c]{13823} &  \makecell[c]{13222}\\  
\makecell[c]{ICEWS2018} &  \makecell[c]{23033} &  \makecell[c]{256} & \makecell[c]{304} &  \makecell[c]{24 hours}&  \makecell[c]{373018} &  \makecell[c]{45995} &  \makecell[c]{49545}\\ 
\makecell[c]{WIKI} &  \makecell[c]{12554} &  \makecell[c]{24} & \makecell[c]{232}  &  \makecell[c]{1 year}&  \makecell[c]{539286} &  \makecell[c]{67538} & \makecell[c]{63110}\\  
 \makecell[c]{YAGO} &  \makecell[c]{10623} & \makecell[c]{10} & \makecell[c]{189} & \makecell[c]{1 year}& \makecell[c]{161540} &  \makecell[c]{19523} &  \makecell[c]{20026}\\  
\end{tabular}}
\caption{Dataset statistics.} 
\label{tab:datasets}  
\end{table}

\begin{table}[H]
\centering
\small
\setlength\tabcolsep{6pt}
\begin{tabular}{llllll}
\hline
Datasets &  \makecell{$N_{\emph{entity}}^{\emph{unseen}}$} & \makecell{$N_{\emph{quad}}^{\emph{unseen~entity}}$} &  \makecell{$N_{\emph{quad}}^{\emph{unseen~object}}$} & \makecell{$N_{\emph{quad}}^{\emph{unseen~subject}}$} & \makecell{$N_{\emph{quad}}^{\emph{unseen~subject\&object}}$}\\ 
\hline 
\makecell{ICEWS2014}  & \makecell{496} & \makecell{862 (6.52\%)} & \makecell{438}& \makecell{497} & \makecell{73} \\  
\makecell{ICEWS2018}  & \makecell{1140} & \makecell{1948 (3.93\%)} & \makecell{975}& \makecell{1050} & \makecell{77}\\   
\makecell{WIKI }  & \makecell{2968} & \makecell{27079 (42.91\%)} & \makecell{11086} & \makecell{22967} & \makecell{6974}\\ 
\makecell{YAGO}  & \makecell{540} & \makecell{1609 (8.03\%)} & \makecell{1102}& \makecell{873} & \makecell{366}\\
\hline
\end{tabular}
\caption{Number of unseen entities and quadruples containing unseen entities in the test set.}
\label{tab:dataset2}
\end{table}

\subsection{Datasets}
We assess our model performance on four public TKGs datasets : ICEWS14~\cite{ICEWS}, ICEWS18~\cite{ICEWS}, WIKI~\cite{WIKI}, YAGO~\cite{YAGO}. Therein, ICEWS14 and ICEWS18 are the two subsets of the Integrated Crisis Early Warning System (ICEWS) dataset, they contain the events in ICEWS that occurred in 2014 and 2018, respectively.
WIKI primarily extracts the temporal events from the Wikipedia dataset, and the temporal facts in YAGO mainly come from Wikipedias, WordNet, and GeoNames.
The dataset is divided by timestamps in the form :  $\emph{train~time} < \emph{valid~time} < \emph{test~time}$, and more data information is summarized in Table~\ref{tab:datasets} and Table~\ref{tab:dataset2}. Note that, in Table~\ref{tab:dataset2}, $N_{\emph{entity}}^{\emph{unseen}}$ denotes the number of new entities in the test set. $N_{\emph{quad}}^{\emph{unseen~entity}}$ represents the number of quadruples that the entities are unseen. $N_{\emph{quad}}^{\emph{unseen~object}}$ is the number of quadruples that object entities are unseen. $N_{\emph{quad}}^{\emph{unseen~subject}}$ denotes the number of quadruples that subject entities are unseen. $N_{\emph{quad}}^{\emph{unseen~subject\&object}}$ represents the number of quadruples that both subject entities and object entities are unseen.

\subsection{Evaluation metrics}
We conduct the temporal knowledge prediction task for evaluating our model. To be specific, we predict two type of queries in the test set : $q_o = (s_q, r_q, ?, t_q)$ and $q_s = (?, r_q, o_q, t_q)$. We add inverse facts into the training dataset, and transform predicting the subject entity for $(?, r_q, o_q, t_q)$  into predicting the object entity for $(o_q, r_q^{-1}, ?, t_q)$ without loss of generality. Given the ground-truth $o_{q}$ and $s_{q}$, we use Hits$@$1/3/10 and MRR~\cite{MRR} to assess our model performance. MRR is the average of the reciprocal of the mean rank (MR) assigned to the true triple overall candidate triples, which is defined as follows :
\begin{eqnarray}
\begin{aligned}
\label{eq::MRR}
\emph{MRR}=\frac{1}{2 \ast \vert test\vert}\sum_{q \in test}\!(\frac{1}{\emph{rank}(o_q|q_o)} + \frac{1}{\emph{rank}(s_q|q_s)})
\end{aligned} 
\end{eqnarray}
Hits@$n$ measures the percentage of test set rankings where a true triple is ranked within the top $n$ candidate triples, defined as:
\begin{eqnarray}
\begin{aligned}
\label{eq::hits}
\emph{Hits@n} = \frac{1}{2 \ast \vert test\vert} \sum_{q \in test} (\mathbb{I}\{\emph{rank}(o_q|q_o)\} \le n 
+ \mathbb{I}\{\emph{rank}(s_q|q_s)\}\le n)
\end{aligned} 
\end{eqnarray}

Furthermore, in this work, we employ the time-aware filtering~\cite{XERTE} scheme instead of the static filtering setting to filter out the quadruples that are genuine at the timestamp, thus obtain more reasonable results. 
\subsection{Baselines}
To assess the proposed model performance comprehensively, we compare it against static KGs reasoning models including TransE~\cite{TransE}, DistMult~\cite{DisMult}, ComplEx~\cite{ComplEx}, MINERVA~\cite{MINERVA}, interpolation TKGs reaoning models including  TTransE~\cite{tTransE}, TA-DistMult~\cite{TADistMult}, DE-SimplE~\cite{DESimplE}, TNTComplEx~\cite{TNTComplex}, and several extrapolation TKGs reasoning models including RE-NET~\cite{Renet}, CyGNet~\cite{CyGNet}, TANGO~\cite{TANGO}, XERTE~\cite{XERTE}, and TITer~\cite{TITer}.

In addition, to evaluate the importance of different components of our model, we propose several variants of our model by adjusting the use of its components.
(1) policy network I: we only use policy network I to obtain the final candidate action score. (2) policy network II: we only use policy network II. (3) w/o TRE: `w/o' refers to `without', TRE denotes temporal relation encoder.  (4) w/o GM: GM represents the gating mechanism, and we use the mean results of policy network I and policy network II as the  final candidate action score. (5) w/o Semantic Edges: It means removing the proposed semantic edges in TKGs.

\subsection{Implementation Details}
In our experiments, we conduct the proposed model in Pytorch~\cite{pytorch} framework. The model is optimized by Adam~\cite{Adam} algorithm with the batch size of 512 and the learning rate of 0.001.  We set the entity embedding dimension to 100, the relation embedding dimension to 100, the timestamp embedding dimension to 100, and the discount factor of REINFORCE algorithm  to 0.95. In addition, at each hop,  for all the candidate actions of the given state, we need to sample $N$ latest candidate actions to score. Here, $N$ is 100 on ICEWS14 and ICEWS18 dataset, 90 on WIKI, 100 on YAGO, and the length of the search path $K$ is 3. In temporal relation encoder, the activation function $\sigma$ is $tanh$. Note that, we reproduce the results of MINERVA with $N=50$ on all datasets, the results of some baseline models are taken from the reported results of TANGO*~\cite{TANGO2}, XERTE~\cite{XERTE}, and TITer~\cite{TITer}.

\renewcommand{\arraystretch}{1.5} 
\setlength\tabcolsep{1.6pt}
\begin{table}[H]
\hspace*{-0.4cm}
\centering
\fontsize{6}{6}\selectfont
\begin{threeparttable}
\begin{tabular}{l|llll|llll|llll|llll}
\toprule
\multirow{4}{*}{Method}&
\multicolumn{4}{c|}{\bf ICEWS2014}&\multicolumn{4}{c|}{\bf ICEWS2018}&\multicolumn{4}{c|}{\bf WIKI}&\multicolumn{4}{c}{\bf YAGO}\cr
\cmidrule(lr){2-5} \cmidrule(lr){6-9}\cmidrule(lr){10-13}\cmidrule(lr){14-17}
&MRR&Hit@1&Hit@3&Hit@10&MRR&Hit@1&Hit@3&Hit@10&MRR&Hit@1&Hit@3&Hit@10&MRR&Hit@1&Hit@3&Hit@10\cr
\midrule
TransE&0.224&0.133&0.256&0.412&0.124&0.058&0.128&0.251&-&-&-&-&-&-&-&-\cr
DistMult&0.276&0.181&0.311&0.469&0.101&0.045&0.103&0.212&0.496&0.461&0.528&0.541&0.548&0.473&0.598&0.685\cr
ComplEx&0.308&0.215&0.344&0.495&0.210&0.118&0.234&0.398&-&-&-&-&-&-&-&-\cr
MINERVA*&0.322&0.254&0.363&0.475&0.210&0.153&0.275&0.330&0.572&0.540&0.593&0.604&0.759&0.727&0.768&0.783\cr
\hline
T-TransE&0.134&0.031&0.173&0.345&0.083&0.019&0.085&0.218&0.292&0.216&0.344&0.423&0.311&0.181&0.409&0.512\cr
TA-DistMult&0.264&0.170&0.302&0.454&0.167&0.086&0.184&0.335&0.445&0.399&0.487&0.517&0.549&0.481&0.596&0.667\cr
De-simplE&0.326&0.244&0.356&0.491&0.193&0.115&0.218&0.348&0.454&0.426&0.477&0.495&0.549&0.516&0.573&0.601\cr
TNTComplEx&0.321&0.233&0.360&0.491&0.212&0.132&0.240&0.369&0.450&0.400&0.493&0.520&0.579&0.529&0.613&0.666\cr
\hline
RE-NET&0.382&0.286&0.413&0.545&0.288&0.190&0.324&0.475&0.496&0.468&0.511&0.534&0.580&0.530&0.610&0.662\cr
CyGNet&0.327&0.236&0.363&0.506&0.249&0.159&0.282&0.426&0.338&0.290&0.361&0.418&0.520&0.453&0.561&0.637\cr
TANGO-Tucker&-&-&-&-&0.286&0.193&0.321&0.470&0.504&0.485&0.514&0.535&0.578&0.530&0.607&0.658\cr
TANGO-DistMult&-&-&-&-&0.267&0.179&0.300&0.440&0.511&0.496&0.521&0.533&0.627&0.591&0.603&0.679\cr
XERTE&0.407&0.327&0.456&0.573&0.293&0.210&0.335&0.464&0.711&0.680&0.761&0.790&0.841&0.800&0.880&0.897\cr
TITer&0.417&0.327&0.464&0.584&0.299&0.220&0.334&0.448&0.755&0.729&0.774&0.790&0.874&0.848&0.899&0.902\cr
\hline
{\bf Ours}&{\bf0.429}&{\bf0.348}&{\bf0.490}&{\bf0.608}&{\bf0.313}&{\bf0.229}&{\bf0.349}&{\bf0.476}&{\bf0.777}&{\bf0.756}&{\bf0.795}&{\bf0.804}&{\bf0.894}&{\bf0.865}&{\bf0.924}&{\bf0.926}\cr
\bottomrule
\end{tabular}
\caption{The results of link prediction on ICEWS14, ICEWS18, WIKI, and YAGO datasets. Compared metrics are filtered MRR and Hits@1/3/10. Best results are in bold.}
\label{tab:raw_performance_comparison} 
\end{threeparttable}
\end{table}

\subsection{Results and Analysis}
\subsubsection{Comparative Study}

Table~\ref{tab:raw_performance_comparison} presents the temporal knowledge prediction results of the proposed model and baseline models on four TKG datasets, and we have the following observations and analyses from the results. First, intuitively, our model outperforms all baseline models on four datasets, which justifies that our method is feasible and achieves considerable results. Second, especially, compared with the related baseline model TITer, our model achieves $2.2\%$ and $2.7\%$ improvement in MRR and Hits@1 on the WIKI dataset, respectively, and obtains different degrees of improvement on the other three datasets. Third, the static reasoning model TransE is superior to its extension version T-TransE on ICEWS2014 dataset, the main reason causes this phenomenon is that TransE and T-TransE are designed for static and interpolation TKGs reasoning, respectively, and they all can not model the unseen future timestamps in query, thus the training time information may affect the prediction performance of T-TransE. 

Fourth, the performance of MINERVA is much higher than other static models, all interpolation TKGs models, and part of extrapolation TKGs models on WIKI and YAGO dataset, this is mainly because most entities in WIKI and YAGO have a small number of neighbouring entities, which allows reinforcement learning-based neighbour search algorithms to find answer entities quickly and accurately. This also is one of the reasons why our method achieves considerable performance on these two datasets.

\subsubsection{Ablation Study}

To verify the effectiveness of different components of the proposed model, we present the results of five variants on ICEWS14 in Table~\ref{tab:AblationStudy}. Policy Network I and II exert different functions in our model, and each of them can calculate the probability distribution of candidate actions and predict the answer for the query. From the results, we can obtain that our model outperforms component models Policy Network I and II, which shows that our model can integrate these two component models adaptively by using the gating mechanism and achieve remarkable results. In addition, we also observe that if we sum the outputs of the two sub-policy networks directly instead of using the gating mechanism, which leads to a drop of $3.5\%$ on MRR, indicating that the gating mechanism can help our model to adaptively focus on the destination answer of the query. The temporal relation encoder is a crucial component in our model, to assess it, we remove  the temporal relation encoder module in policy network II and adopt the same encoding method as policy network I, the model performance is dropped by $3.9\%$ on MRR compared with the proposed model. Moreover, we propose to add semantic edges into TKGs to help the agent find the actions in the case that the query contains an unseen entity, thus we remove the semantic edges here to evaluate their importance. From the results, we can obtain that this variant model performance is dropped by $1.1\%$ on MRR compared with the proposed model. This also demonstrates the effectiveness of  the temporal relation encoder in temporal knowledge representation.
\begin{table}
\centering
\fontsize{9}{9}\selectfont
\begin{tabular}{lllllll}
\hline
Model variant &  \makecell{MRR} & \makecell{Hits@1} &  \makecell{Hits@3} & \makecell{Hits@10}  \\ 
\hline 
\makecell{Policy Network I}  & \makecell{0.391} & \makecell{0.290} & \makecell{0.441}& \makecell{0.590} \\  
\makecell{Policy Network II}  & \makecell{0.415} & \makecell{0.328} & \makecell{0.460}& \makecell{0.578} \\   
\makecell{w/o TRE }  & \makecell{0.390} & \makecell{0.288} & \makecell{0.441}& \makecell{0.593} \\ 
\makecell{w/o GM}  & \makecell{0.394} & \makecell{0.295} & \makecell{0.444} & \makecell{0.588}\\
\makecell{w/o Semantic Edges}  & \makecell{0.418} & \makecell{0.336} & \makecell{0.475} & \makecell{0.592}\\
\makecell{\bf Ours}  & \makecell{\bf0.429} & \makecell{\bf0.348} & \makecell{\bf0.490}& \makecell{\bf0.608} \\ 
\hline
\end{tabular}
\caption{Ablation study on ICEWS14 dataset.}
\label{tab:AblationStudy}
\end{table}
\subsubsection{Ablation Study II}
Our motivation for designing such a novel model is to better handle the two cases where the query contains seen and unseen entities in a unified framework. Specifically, when the query contains seen entity, the policy network I and II can capture the evolutionary patterns of the facts cooperatively. And when the query contains the unseen entity, the policy network II is able to exert its particular function. Therefore, in this section, we justify our design from the two perspectives.
\begin{figure}[H]
\hspace*{-1.2cm}
\setlength{\abovecaptionskip}{-0.4cm}
\setlength{\belowcaptionskip}{-0.cm}  
\centering
\includegraphics[height=6.3cm, width=12cm]{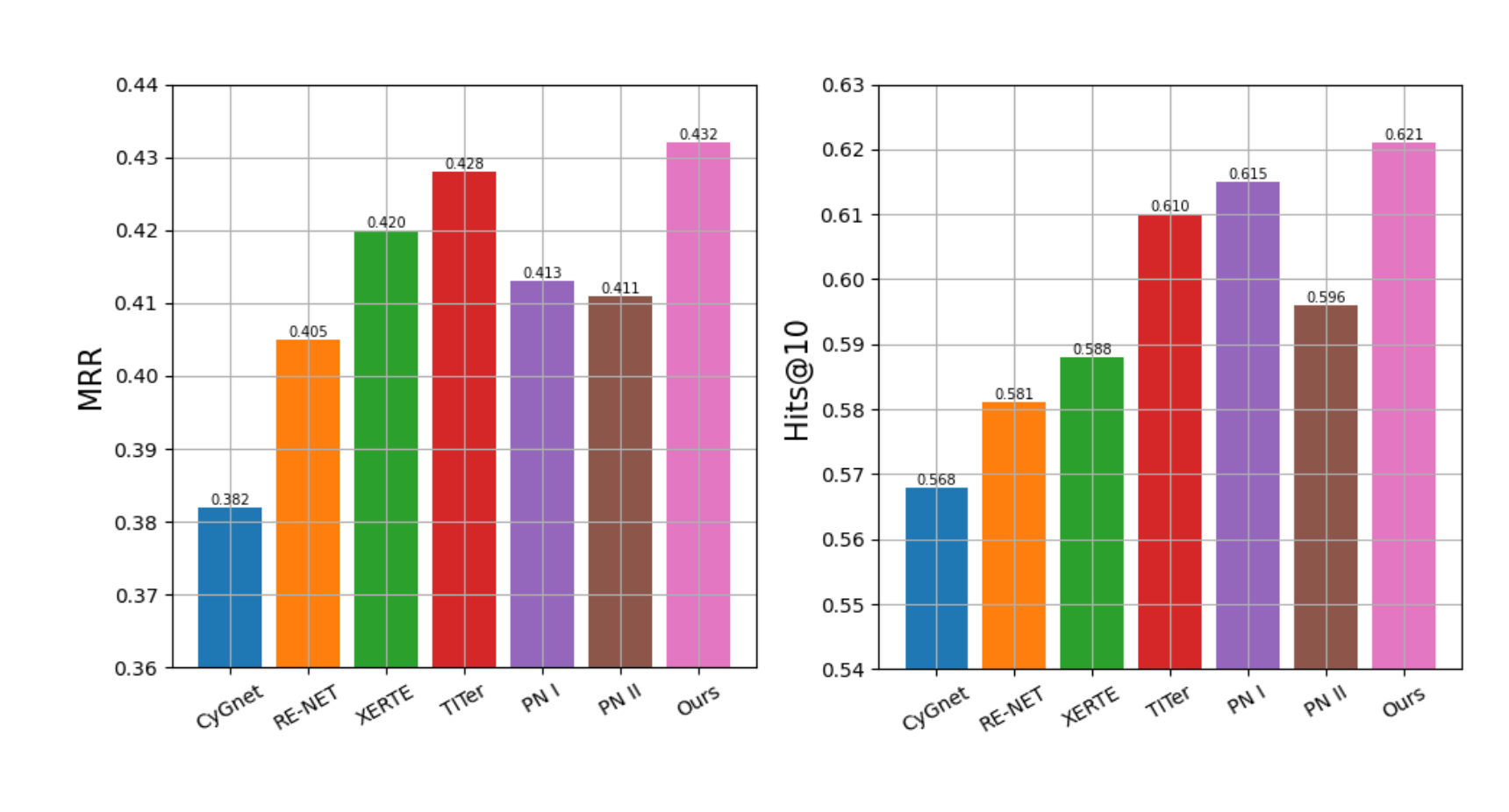} 
\caption{Temporal knowledge prediction results on the subset of ICEWS14 that contain seen entities. PN I and PN II represent Policy Network I and Policy Network II, respectively.}
\label{fig::Motivation1}
\end{figure}

We first verify our design from the first perspective that the policy network I and II can capture the evolutionary patterns of the facts cooperatively when the query contains seen entity, thus we conduct the link prediction on the subset of ICEWS14 that contain seen entities, and the results are presented in Figure~\ref{fig::Motivation1}. From the results, we can obtain that our model is superior to the baseline models. Most importantly, the proposed model outperforms the component models policy network I and II. 
\begin{figure}[H]
\hspace*{-1.2cm}
\setlength{\abovecaptionskip}{-0.4cm}
\setlength{\belowcaptionskip}{-0.cm}  
\centering
\includegraphics[height=6.3cm, width=12cm]{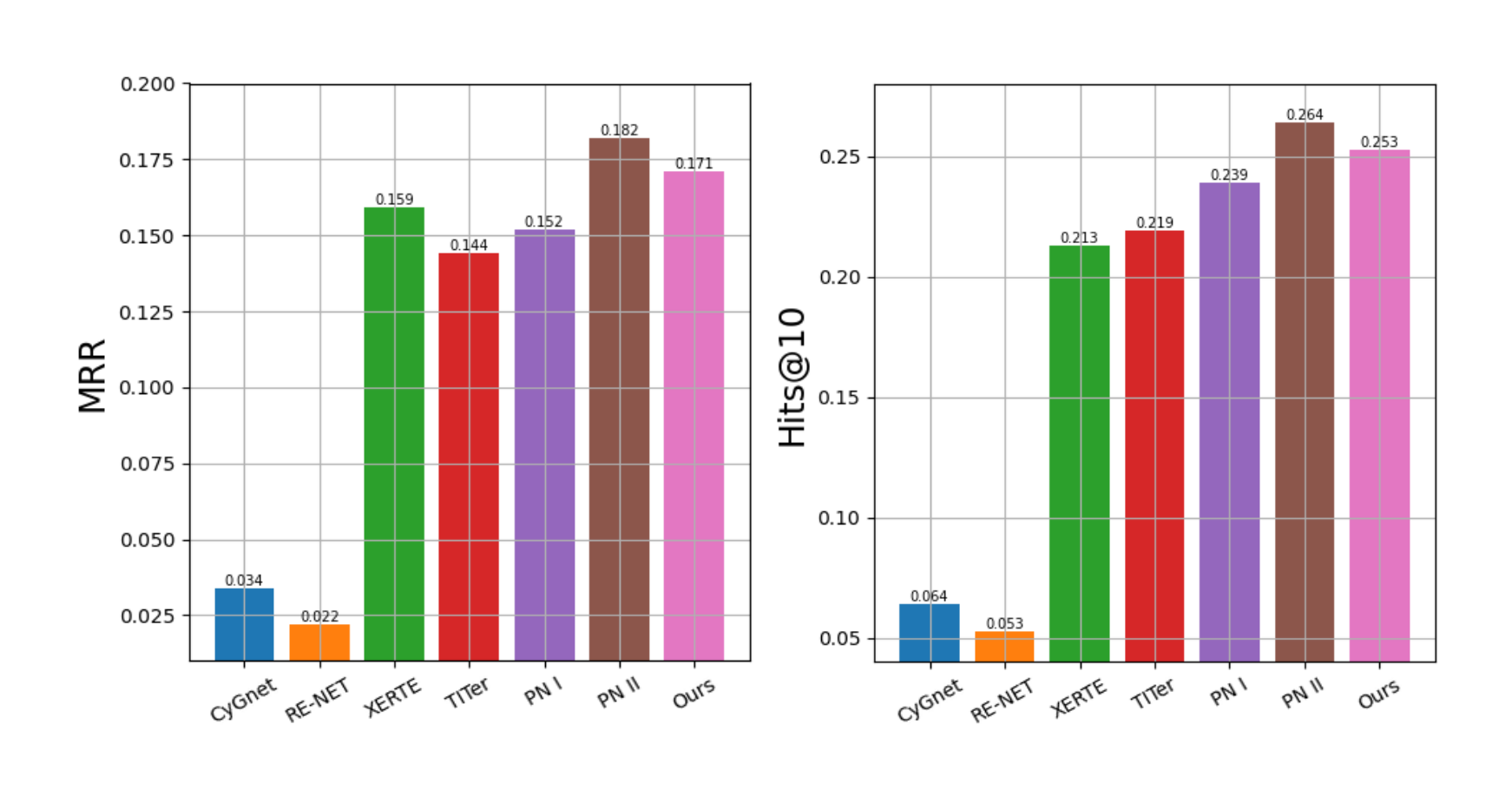} 
\caption{Temporal knowledge prediction results on the subset of ICEWS14 that contain unseen entities. PN I and PN II represent Policy Network I and Policy Network II, respectively.}
\label{fig::Motivation2}
\end{figure}

To justify our design from the second view that the policy network II can exert its particular function in the case where the query contains the unseen entity, we also conduct the link prediction on the subset of ICEWS14 that contain unseen entities, and the results are shown in Figure~\ref{fig::Motivation2}. Intuitively, we can obtain that policy network II has explicit advantages in handling the case that the query contains an unseen entity from the results, and our model can absorb the advantages of policy network II to achieve considerable performance. The above results demonstrate the effectiveness of our design and provide a shred of evidence that our model can predict events in complex scenarios. 
\begin{table}[H]
\centering
\small
\setlength\tabcolsep{6pt}
\begin{tabular}{llll}
\hline
\hline
 \makecell{Query} \!\!\!&\!\!\!  \makecell{Path} \!\!\!&\!\!\! \makecell{Answer}\\ 
\hline
 \makecell{(\emph{Ed Royce}, \emph{Make pessimistic} \\\emph{comment}, ?, \emph{2014-11-23})$^{\dagger}$} & \makecell{(\emph{Ed Royce}, \emph{Criticize or denounce}, \\\emph{Nuri al-Maliki}, \emph{2014-2-6}) $\Rightarrow$
(\emph{Nuri al-Maliki},\\ \emph{Make statement}, \emph{Iran}, \emph{2014-1-25}) $\Rightarrow$
(\emph{Iran},\\ \emph{self-loop}, \emph{\textbf {Iran}}, \emph{2014-1-25)}} &  \makecell{\emph{\textbf {Iran}}}\\  
\hline  
\makecell{(\emph{Cabinet USA}, \emph{Consult}, ?,\\ \emph{2014-11-11})$^{\dagger}$} & \makecell{(\emph{Cabinet USA}, \emph{Deny responsibility}$^{-1}$, \\\emph{Sergey}, \emph{2014-11-9}) $\Rightarrow$ (\emph{Sergey}, \emph{self-loop}, \\\emph{Sergey},\emph{2014-11-9}) $\Rightarrow$ (\emph{Sergey}, \emph{self-loop},\\ \emph{\textbf {Sergey}}, \emph{2014-11-9)}} & \makecell{\emph{\textbf {Sergey}}}\\
\hline 
\makecell{(\emph{Mehdi Hasan}, \emph{Make an appeal}\\ \emph{or request}, ?, \emph{2014-11-12})$^{\ddagger}$} & \makecell{(\emph{Shivraj}, \emph{Make an appeal or request},\\ \emph{India Citizen}, \emph{2014-6-22}) $\Rightarrow$ (\emph{India} \\\emph{Citizen}, \emph{self-loop}, \emph{India Citizen},\\ \emph{2014-4-7})
 $\Rightarrow$ (\emph{India Citizen}, \emph{self-loop},\\ \emph{\textbf{India Citizen}}, \emph{2014-6-22})} &  \makecell{\emph{\textbf{India Citizen}}} \\ 
\hline 
 \makecell{(\emph{India Chief}, \emph{Threaten},\\ ?, \emph{2014-11-19})$^{\ddagger}$} 
 & \makecell{(\emph{Australia Citizen}, \emph{Threaten}, \emph{India Citizen}, \\ \emph{2014-7-30})
 $\Rightarrow$ (\emph{India Citizen}, \emph{self-loop}, \\\emph{India Citizen}, \emph{2014-8-20})
 $\Rightarrow$ (\emph{India Citizen}, \\\emph{self-loop}, \emph{\textbf{India Citizen}}, \emph{2014-6-30)} } &  \makecell{\emph{\textbf {India Citizen}}} \\ 
\hline
\hline
\end{tabular}
\caption{Case study for two kinds of queries on ICEWS2014 dataset. $(\cdot)^{\dagger}$ represents the query with seen entity, and $(\cdot)^{\ddagger}$ indicates the query with unseen entity.}
\label{tab:casestudy}
\end{table}

\subsubsection{Case Study}

Table~\ref{tab:casestudy} provides specific reasoning paths of four queries in the test set of ICEWS2014, and the first two of them are the query with a seen entity, and the last two of them are the query with an unseen entity. For the query with seen entity (\emph{Ed Royce}, \emph{Make pessimistic comment}, ?, \emph{2014-11-23}), the agent arrives at the entity $\emph{Nuri al-Maliki}$ according the temporal edge (\emph{Ed Royce}, \emph{Criticize or denounce}, \emph{Nuri al-Maliki}, \emph{2014-2-6}) from the start entity $\emph{Ed Royce}$ in the first step. After three steps, the agent reaches the answer entity $\emph{Iran}$.  While for the query with an unseen entity (\emph{Mehdi Hasan}, \emph{Make an appeal} \emph{or request}, ?, \emph{2014-11-12}), the entity $\emph{Mehdi Hasan}$ is unseen in the historical facts, the agent can not find temporal edges to traverse except for the self-loop edge. In this case, the agent first reaches the seen entity $\emph{India Citizen}$ thorough the semantic edge (\emph{Shivraj}, \emph{Make an appeal or request}, \emph{India Citizen}, \emph{2014-6-22}), then arrives at the answer thorough the temporal edges or self-loop edges in the following two steps. From the above results, we can obtain that the semantic edge bridge the gap between the query with an unseen entity and the historical facts, and it exerts an important function in our method.

\section{Conclusion}
This work proposes a novel reinforcement learning model for temporal knowledge prediction task on TKGs. The model is presented mainly to deal with two problems, one is to model the evolutionary patterns of the facts, and the other is to deal with uncertain queries that the query contains seen or unseen entities. Thus we design a novel adaptive pseudo-siamese policy network to deal with the two problems in a unified framework, policy networks I is used to capture the static evolutionary patterns of the facts, and policy networks II with the designed temporal relation encoder is employed to capture the temporal evolutionary patterns of the facts and deal the special case that the query contains the unseen entity. Finally, we develop a gating mechanism to integrate the results of the two sub-policy networks adaptively to help the agent focus on the destination answer. Experimental results on four public datasets explicitly demonstrate the effectiveness and superiority of the proposed model on the temporal knowledge prediction task.

\section*{References}
\bibliography{TKP}
\end{document}